\documentclass{article}
\usepackage{ijcai21}

\usepackage{times}
\usepackage{soul}
\usepackage{url}
\usepackage[hidelinks]{hyperref}
\usepackage[utf8]{inputenc}
\usepackage[small]{caption}
\usepackage{graphicx}
\usepackage{amsmath}
\usepackage{amsthm}
\usepackage{booktabs}
\usepackage{algorithm}
\usepackage{algorithmic}
\usepackage{blindtext}
\usepackage[inline, shortlabels]{enumitem}
\DeclareMathOperator{\simm}{\textit{sim}}

\DeclareMathOperator{\QW}{\textit{QW}}
\DeclareMathOperator{\CW}{\textit{CW}}

\newcommand{\eg}{e.\,g.}
\newcommand{\wrt}{w.\,r.\,t. }
\newcommand{\ie}{i.\,e.}

\title{Informed Machine Learning for Improved Similarity Assessment in Process-Oriented Case-Based Reasoning}

\author{
Maximilian Hoffmann$^{1,2}$
\and
Ralph Bergmann$^{1,2}$
\affiliations
$^1$Business Information Systems II, University of Trier, 54296 Trier, Germany\\\url{http://www.wi2.uni-trier.de}\\
\{hoffmannm, bergmann\}@uni-trier.de\\
$^2$German Research Center for Artificial Intelligence (DFKI) \\ Branch University of Trier, Behringstraße 21, 54296 Trier, Germany\\
\emails
\{maximilian.hoffmann, ralph.bergmann\}@dfki.de
}

\begin{document}

\maketitle

\begin{abstract}
Currently, \textit{Deep Learning} (DL) components within a \textit{Case-Based Reasoning} (CBR) application often lack the comprehensive integration of available domain knowledge. The trend within machine learning towards so-called \textit{Informed machine learning} can help to overcome this limitation. In this paper, we therefore investigate the potential of integrating domain knowledge into \textit{Graph Neural Networks} (GNNs) that are used for similarity assessment between semantic graphs within process-oriented CBR applications. We integrate knowledge in two ways: First, a special data representation and processing method is used that encodes structural knowledge about the semantic annotations of each graph node and edge. Second, the message-passing component of the GNNs is constrained by knowledge on legal node mappings. The evaluation examines the quality and training time of the extended GNNs, compared to the stock models. The results show that both extensions are capable of providing better quality, shorter training times, or in some configurations both advantages at once.
\end{abstract}

\section{Introduction}
\label{sec:Introduction}

\textit{Case-Based Reasoning} (CBR) \cite{Aamodt.1994_CBR,Richter.2013_CBRTextbook} is used widespread across different domains, \eg, modeling of cooking recipes \cite{Hoffmann.2020_GraphEmbedding}, prediction of seawater temperatures \cite{Corchado.2001_AdaptationNeuralNetworkSupport}, natural language processing of support tickets \cite{Amin.2020_DeepKAF}, and assisted reuse of data mining workflows \cite{Zeyen.2019_AdaptationScientificWorkflows}. One of its strengths is the use of structured domain knowledge that is modeled, among others, for the case representation, for the definition of similarity measures, and for case adaptation methods \cite{Richter.2013_CBRTextbook}. An integral part of recent CBR research is the use of \textit{Deep Learning} (DL) methods, which is indicated by workshops dedicated to this topic, \eg, the workshops on the International Conference on Case-Based Reasoning (ICCBR) in 2017 \cite{SanchezRuiz.2017_DLWorkshop} and 2019 \cite{Kapetanakis.2019_DLWorkshop}, and a variety of published papers, \eg, \cite{Mathisen.2021_Aquaculture,Hoffmann.2020_GraphEmbedding,Amin.2020_DeepKAF,Klein.2019_MACFACEmbedding}. Many of these works are hybrid approaches where DL components are integrated in the underlying CBR methodology \cite{Watson.1999_CBRMethodology} to solve certain core tasks such as similarity assessment or case adaptation. This is a reasonable choice due to the ability of DL to automatically learn patterns from available data, diminishing the need for time-consuming manual data analysis and extraction \cite{Stahl.2006_KnowledgeIntegrationInSearchProcess,Leake.2020_CBRMethodologyToDeepLearning}. However, most of these hybrid approaches lack a comprehensive integration of the CBR-provided knowledge into the DL methods, leading to possibly unused potential of improved quality and performance. A recent trend in artificial intelligence research explicitly deals with such an integration of symbolic knowledge into machine learning methods, \ie, \textit{Informed Machine Learning} \cite{VonRueden.2019_InformedMachineLearning}. The core ideas of informed machine learning can be directly used in our scenario of hybrid approaches by combining the strengths of CBR and DL: On the one hand, CBR offers a large amount of domain knowledge that is often of high quality. On the other hand, the DL components provide the flexibility and expressiveness to integrate and process this domain knowledge.

In this paper, we want to investigate the possibilities of informed machine learning in the research field of \textit{Process-Oriented Case-Based Reasoning} (POCBR) \cite{Minor.2014_POCBR}. POCBR is a subfield of CBR that deals with procedural knowledge such as semantic graphs \cite{Bergmann.2014_NESTGraph}. We examine how POCBR domain knowledge can be used to improve DL-based similarity assessment between semantic graphs. The presented methods extend the DL models from our previous work \cite{Hoffmann.2020_GraphEmbedding} and explain the representation and integration of domain knowledge for two specific scenarios. We also put these two methods into the context of an informed machine learning taxonomy by von Rueden et al. [\citeyear{VonRueden.2019_InformedMachineLearning}] to encourage a broader discussion about this topic in CBR research. The paper is structured as follows: Section \ref{sec:Foundations} shows foundations regarding the used semantic graph representation and the similarity assessment between these graphs and, additionally, related work is presented. Further, Sect.~\ref{sec:Concept} elaborates on our approach for integrating domain knowledge into semantic graph embedding models. To evaluate our approach and to briefly discuss the possibilities of domain knowledge integration for DL methods in CBR, we present an experimental evaluation in Sect.~\ref{sec:Evaluation}. Finally, Sect.~\ref{sec:ConclusionAndFutureWork} concludes the paper and shows areas of future work.

\section{Foundations and Related Work}
\label{sec:Foundations}

The foundations include the semantic workflow representation that we use in our concept and experiments as well as the similarity assessment between pairs of these workflows (see Sect.~\ref{sec:Foundations:subsec:SemanticGraphs}). We also present the core ideas of informed machine learning, as described by von Rueden et al. [\citeyear{VonRueden.2019_InformedMachineLearning}] (see Sect.~\ref{sec:Foundations:subsec:InformedML}). Further, related work is presented (see Sect.~\ref{sec:Foundations:subsec:RelatedWork}).

\subsection{Semantic Workflow Representation and Similarity Assessment} 
\label{sec:Foundations:subsec:SemanticGraphs}

\begin{figure}[htb]
    \centering
    \includegraphics[width=1.0\columnwidth]{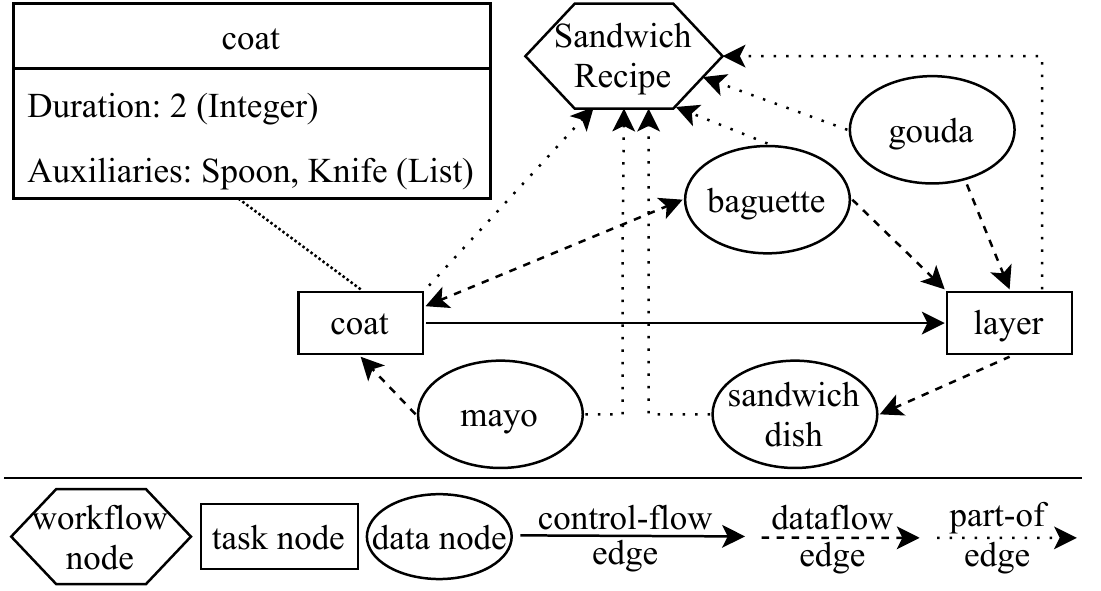}
    \caption[Exemplary Cooking Recipe represented as NEST Graph]{Exemplary Cooking Recipe represented as NEST Graph}
    \label{fig:simpleCookingWorkflowNoSemDescr}
\end{figure}

Our main focus in this paper is similarity assessment between semantic graphs in POCBR. We represent all case bases and queries as semantically annotated directed graphs referred to as \emph{NEST} graphs introduced by Bergmann and Gil [\citeyear{Bergmann.2014_NESTGraph}]. More specifically, a \emph{NEST} graph is a quadruple \(W=(N,E,S,T)\) that is composed of a set of nodes \(N\) and a set of edges \(E \subseteq N \times N\). Each node and each edge has a specific type from \(\Omega\) that is indicated by the function \(T : N \cup E \rightarrow \Omega\). Additionally, the function \(S : N \cup E \rightarrow \Sigma\) assigns a semantic description from \(\Sigma\) (\textit{semantic metadata language}, \eg, an ontology) to nodes and edges. Whereas nodes and edges are used to build the structure of each workflow, types and semantic descriptions are additionally used to model semantic information. Hence, each node and each edge can have a semantic description. Figure \ref{fig:simpleCookingWorkflowNoSemDescr} shows a simple example of a \emph{NEST} graph that represents a cooking recipe for making a sandwich. The mayo-gouda sandwich is prepared by executing the cooking steps \texttt{coat} and \texttt{layer} (task nodes) with the ingredients \texttt{mayo}, \texttt{baguette}, \texttt{sandwich dish}, and \texttt{gouda} (data nodes). All components are linked by edges that indicate relations, \eg, \texttt{mayo} is consumed by \texttt{coat}. Semantic descriptions of task nodes and data nodes are used to further specify semantic information belonging to the workflow components. Figure \ref{fig:simpleCookingWorkflowNoSemDescr} shows an example of the semantic description of the task node \texttt{coat}. The provided information is used to describe the task more precisely. In this case, a spoon and a baguette knife is needed to execute the task (\texttt{Auxiliaries}) and the estimated time that the task takes is two minutes (\texttt{Duration}).

We use the CBR framework ProCAKE \cite{Bergmann.2019_ProCAKE} to process NEST graphs, thus their semantic descriptions can be represented in various ways. As seen in the example, atomic data types such as strings and numerics can be combined within composite data types such as attribute-value pairs and lists. This complexity and flexibility in representation is not easy to handle for DL-based approaches (see \cite{Leake.2020_CBRMethodologyToDeepLearning} and \cite{Hoffmann.2020_GraphEmbedding} for more details) and requires specific data encoding methods which will be a topic in our approach (see Sect.~\ref{sec:Concept}).

Determining the similarity between two \emph{NEST} graphs, \ie, a query workflow $\QW$ and a case workflow $\CW$, requires a similarity measure that assesses the link structure of nodes and edges as well as the semantic descriptions and types of these components. Bergmann and Gil [\citeyear{Bergmann.2014_NESTGraph}] propose a semantic similarity measure that determines a similarity based on the local-global principle. A global similarity, \ie, the similarity between two graphs, is composed of local similarities, \ie, the pairwise similarities of nodes and edges. The similarity between two nodes with identical types is defined as the similarity of the semantic descriptions of these nodes. The similarity between two edges with identical types does not only consider the similarity of the semantic descriptions of the edges, but in addition the similarity of the connected nodes as well. In order to put together a global similarity by aggregating local similarities, the domain's similarity model has to define similarity measures for all components of the semantic description, \ie, \(\simm_\Sigma : \Sigma \times \Sigma \rightarrow [0,1]\). The global similarity of the two workflows \(\simm(\QW, \CW)\) is finally calculated by finding an injective partial mapping \(m\) that maximizes \(\simm_m(\QW, \CW)\).
\begin{multline}
  \simm(\QW,\CW) = max \{ \simm_{m}(\QW,\CW)~|\\\text{admissible mapping}~m \}  
\end{multline}
The complex process of finding a mapping that maximizes the global similarity between a query $\QW$ and a single case $\CW$ is tackled by utilizing an A\textsuperscript{*} search algorithm (see \cite{Bergmann.2014_NESTGraph} for more details). However, A* search is usually time-consuming and can lead to long retrieval times \cite{Zeyen.2020_AStar,Klein.2019_MACFACEmbedding,Hoffmann.2020_GraphEmbedding} which also motivates this work.

\subsection{Informed Machine Learning}
\label{sec:Foundations:subsec:InformedML}
 
von Rueden et al. [\citeyear{VonRueden.2019_InformedMachineLearning}] characterize the integration of prior knowledge, \ie, already existing knowledge from the present domain, into a machine learning pipeline according to three key factors. All three of these key factors further contain several sub-factors. The \textit{source} of the prior knowledge is the first key factor. It is further distinguished between sources with different levels of validation, \eg, knowledge from natural sciences is explicitly validated while common world knowledge is usually not. Besides the source of knowledge, knowing and analyzing its \textit{representation} is also crucial for classifying informed machine learning methods. A knowledge representation can vary from logic rules over knowledge graphs to human feedback, all requiring individual processing strategies. The third factor is the point of knowledge \textit{integration} within the machine learning pipeline. This refers to four possible integration hooks: the training data, the DL architecture (referred to as "hypothesis set" in the original paper \cite{VonRueden.2019_InformedMachineLearning}), the learning algorithm, and the final hypothesis. In our approach, we will particularly focus on the representation and the point of integration of the domain knowledge (see Sect.~\ref{sec:Concept}).

\subsection{Related Work}
\label{sec:Foundations:subsec:RelatedWork}

In CBR research, several approaches have been proposed that use DL methods as a key component. However, the presentation of these approaches is mostly outcome-focused, concentrating on the integration of the DL methods within a CBR application rather than on the usage of CBR-provided knowledge. One of the few examples of the latter approaches is the work of Stahl and Gabel [\citeyear{Stahl.2006_KnowledgeIntegrationInSearchProcess}] where they provide a general learning approach for similarity measures. They identify three different knowledge sources, \ie, similarity meta knowledge, knowledge from the case base, and expert knowledge, that are beneficial for being integrated into their machine learning pipeline. Outcome-focused approaches are much more common. For instance, Corchado and Lees [\citeyear{Corchado.2001_AdaptationNeuralNetworkSupport}] integrate a neural network into CBR retrieval and reuse phases where the network is dynamically retrained during runtime \wrt the current query. The application is used to predict water temperatures along a sea route. Dieterle and Bergmann [\citeyear{Dieterle.2014_ANNInternetDomains}] use neural networks for several tasks within their CBR application that predicts prices of domain names, \eg, for feature-weighting of case attributes. In our previous work \cite{Hoffmann.2020_GraphEmbedding,Klein.2019_MACFACEmbedding} and in other approaches \cite{Mathisen.2021_Aquaculture,Amin.2020_WordEmbeddings}, siamese neural networks are used for learning similarity functions between cases. The resulting similarity functions are mainly used for retrieval situations. The application of DL methods on case adaptation is discussed by several other approaches, \eg, \cite{Leake.2021_AdaptationNeuralNetworks,Liao.2018_MLCaseAdaptation}. These two exemplary approaches pursue the idea of using neural networks in combination with the case-difference heuristic for solving an upcoming problem with known solutions from the case base. DL is also used in the context of textual CBR \cite{Amin.2020_DeepKAF,Amin.2020_WordEmbeddings}. These approaches use siamese neural networks and word embeddings for tasks of natural language processing. Furthermore, Gabel and Godehardt [\citeyear{Gabel.2015_SimilarityClouds}] and Keane and Kenny [\citeyear{Keane.2019_XAI}] tackle a major drawback of DL applications when compared to CBR applications: the reduced explainability. The former approach addresses the problem by projecting DL predictions onto real cases before using them. The latter approach tries to explain the predictions of DL methods with CBR components in a twin-systems approach. To guide research of CBR and DL, Leake and Crandall [\citeyear{Leake.2020_CBRMethodologyToDeepLearning}] elaborate ways where the CBR methodology \cite{Watson.1999_CBRMethodology} can advance DL. They present benefits and drawbacks of both methods and, additionally, identify questions that arise when developing hybrid approaches. These questions further motivate our work since we see knowledge integration along with suitable case representation and similarity assessment, \ie, the main contributions of this paper, as a key factor for answering these questions.

\section{Similarity Assessment with Informed Machine Learning Methods}
\label{sec:Concept}

This section presents our approach for integrating domain knowledge into the \textit{Graph Neural Networks} (GNNs) for POCBR, which were originally proposed in our previous work \cite{Hoffmann.2020_GraphEmbedding}. We present two novel extensions to the GNNs, aimed at improving the similarity assessment capabilities. The extensions are motivated by utilizing existing knowledge present in the underlying CBR application regarding case representation (see Sect.~\ref{sec:Concept:subsec:Extension1}) and similarity matching constraints (see Sect.~\ref{sec:Concept:subsec:Extension2}). Before explaining these methods, we will first give a brief overview of the GNNs that build the foundation for the approach in this paper (see Sect.~\ref{sec:Concept:subsec:GNN}).

\subsection{Graph Neural Networks for Similarity Assessment}
\label{sec:Concept:subsec:GNN}

\begin{figure}[htb]
    \centering
    \includegraphics[width=1\columnwidth]{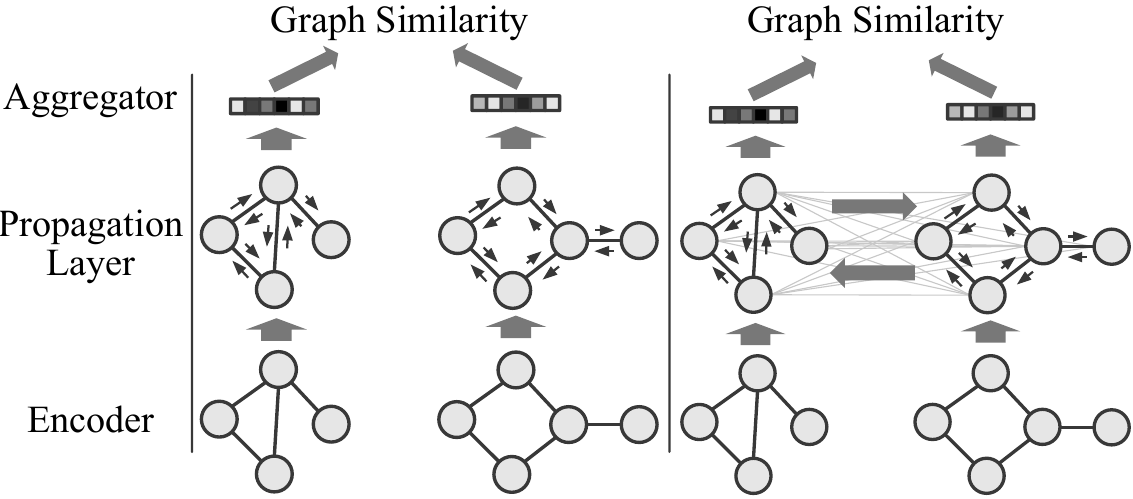}
    \caption[]{GEM (left) and GMN (right) (based on \cite{Li.2019_GraphMatchingNetworks})}
    \label{fig:GEMGMN}
\end{figure}

Two siamese GNNs are used for learning pairwise graph similarities, \ie, the \textit{Graph Embedding Model} (GEM) and the \textit{Graph Matching Network} (GMN) (see Fig. \ref{fig:GEMGMN}). Both models follow the same basic structure but show different levels of expressiveness which leads to different areas of application. The four main components are the \textit{encoder}, the \textit{propagation layer}, the \textit{aggregator}, and the \textit{graph similarity}. The encoder initially performs an encoding and embedding of all nodes and edges from both graphs to form single embedding vectors which further represent individual nodes and edges, respectively. During propagation, the vector-based information is iteratively merged according to the edge structure of the graphs. Thereby, node embedding vectors are updated via element-wise summations according to the information of all incoming edges and the nodes that are connected via these edges. After propagating information for a number of rounds, the aggregator combines the embeddings of all nodes from each graph to form a single whole-graph embedding. The final component forms a single scalar similarity value with the embedding vectors of both graphs.

The difference between both models comes from a different propagation strategy and a different computation of the final similarity. The GEM uses an isolated propagation strategy that only propagates information within a single graph. The aggregated vectors of both graphs are eventually put together to a similarity value via a simple cosine similarity. In contrast, the GMN uses a cross-graph matching component that propagates information across both graphs in an early state of similarity assessment. Thereby, the embedding vector of each node from one graph is compared with the embedding vector of each node from the other graph by means of a vector similarity measure. The higher the similarity and therefore the attention, the stronger is the information flow between two nodes. The whole-graph embedding vectors are eventually fed into a \textit{Multi-Layer Perceptron} (MLP) that generates the final similarity. These differences of both models lead to different reasonable application areas. That is, GEM is faster but less expressive and therefore more suitable to be used as a similarity measure for candidate filtering such as in MAC/FAC applications \cite{Forbus.1995_MACFAC,KendallMorwick.2014_TwoPhaseRetrieval}. GMN is slower than GEM but shows great potential in approximating graph similarities with low margins of error (see the evaluation in \cite{Hoffmann.2020_GraphEmbedding} for more details).

\subsection{Extension 1 - Tree Encoding}
\label{sec:Concept:subsec:Extension1}

The first extension concerns the encoding and processing of semantic descriptions (see Sect.~\ref{sec:Foundations:subsec:SemanticGraphs}) in GEM and GMN. Currently, each atomic entry of a semantic description (\eg, integer, float, string) is encoded as a single vector consisting of a one-hot type encoding part with a proprietary data encoding part specific to each data type. Composite data types (\eg, attribute-value lists, sets) are handled as a collection of atomic encodings which is transformed to a single matrix structure per encoded semantic description. For instance, the exemplary semantic description from Fig. \ref{fig:simpleCookingWorkflowNoSemDescr} shows two composite types, \ie, the attribute-value list \texttt{coat} and the list \texttt{Auxiliaries}, as well as three atomic types, \ie, the duration and both list entries. The encoding of this semantic description is a single matrix consisting of three entries which equals the number of all atomic types. The sequence results from the two composite types that are concatenated.

\begin{figure}[htb]
    \centering
    \includegraphics[width=0.95\columnwidth]{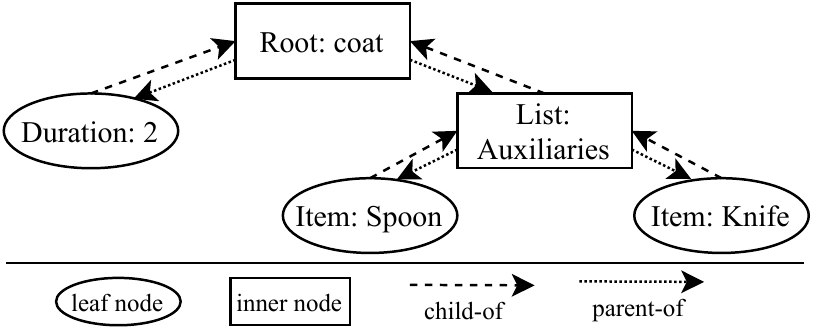}
    \caption[]{Exemplary Tree Structure of an Encoded Semantic Description}
    \label{fig:TreeEncoding}
\end{figure}

In order to process this data in the encoder step of GEM and GMN, the matrix is processed by \textit{Recurrent Neural Networks} (RNNs) which aggregate the sequence of vectors to a single vector. This is not ideal since the underlying sequence structure does not accurately represent the relations of the components of the semantic descriptions. Therefore, with the first extension, we represent the encoded semantic descriptions as \textit{tree structures} and extend the encoders of GEM and GMN to process these tree structures with GNNs. Fig. \ref{fig:TreeEncoding} shows the new format where atomic types are represented as ellipses and composite types as rectangles. The connections between nodes represent their relations, \ie, child-of and parent-of. Regarding the computations in the neural network, we handle this representation as a graph that is processed with a GNN \cite{Battaglia.2018_RelationalInductiveBiases}. Analogously to the general embedding process of GEM and GMN (see Fig. \ref{sec:Concept:subsec:GNN}), information is propagated between the nodes of the graph along the edge connections. The edges are either of type child-of or parent-of, which is given as a piece of information that is propagated. The result after several iterations of propagation is a single embedding vector for each node and each edge of the workflow graph. The used GNN can also be parameterized to either reuse the same neural network layers for each iteration or to use different layers. The latter configuration is more expressive and computationally expensive since parameter learning can be performed individually for each iteration (see both configurations as part of our evaluation in Sect.~\ref{sec:Evaluation}). Regarding the taxonomy of informed machine learning (see Sect.~\ref{sec:Foundations:subsec:InformedML}), this method focuses on knowledge representation and integration. We use a new representation for the semantic descriptions which is very similar to their domain-defined tree structure. We also integrate this representation into the training data and the hypothesis set, in order to enable the model to use the data properly. The source of the data is given by the structural relations of the semantic descriptions. In Sect.~4 of their paper, Leake and Crandall [\citeyear{Leake.2020_CBRMethodologyToDeepLearning}] raise a few key questions for CBR research of DL integrations, with one of those questions concerning case representation. We want to point out that the aforementioned extension contributes to this question. Our approach can also be used as a standalone method for representing and processing structured, object-oriented cases such as the semantic descriptions.

\subsection{Extension 2 - Matching Constraints}
\label{sec:Concept:subsec:Extension2}

The second extension concerns the propagation component of GMN. It is inspired by the graph matching algorithm that we use to assess the similarity of graphs (see Sect.~\ref{sec:Foundations:subsec:SemanticGraphs}). In the propagation component of GMN (see Fig. \ref{fig:GEMGMN}), each node of one graph is compared with each node of the other graph by means of their respective cosine vector similarity (ranged between 0 and 1). The resulting similarity values are activated with a softmax activation in order to generate attention values. This way, information is propagated between the nodes of two graphs \wrt to their pairwise attention.

\begin{figure}[htb]
    \centering
    \includegraphics[width=0.95\columnwidth]{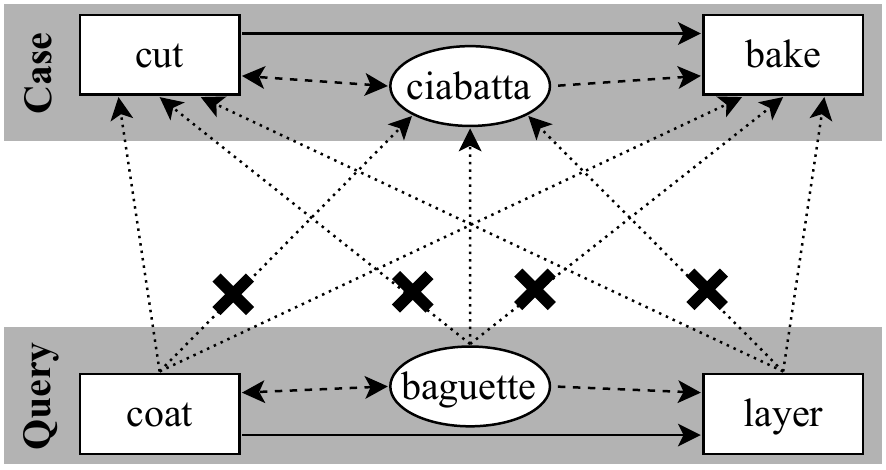}
    \caption[]{Exemplary Matching Process}
    \label{fig:Matching}
\end{figure}

According to the definition of the used graph matching algorithm (see Sect.~\ref{sec:Foundations:subsec:SemanticGraphs}), only nodes and edges with the same type are allowed to be matched. That is, the matching process is constrained according to the types of nodes and edges. These \textit{constraints} can be integrated into the propagation component of GMN by only allowing to determine cosine similarity for nodes of the same type. Fig. \ref{fig:Matching} shows an exemplary matching process of a query and a case graph (grey background) with dotted lines depicting all possible mappings of the query nodes to the case nodes. All illegal mappings are marked with a bold cross. For instance, \texttt{coat} is a task node and can only be matched with other task nodes, leading to an illegal match between \texttt{coat} and \texttt{ciabatta}. The data node \texttt{baguette} has only one possible mapping partner since the case graph only contains one data node, \ie, \texttt{ciabatta}. All pairs of nodes with different types are assigned to a similarity of 0, representing maximum dissimilarity. Thus, their attention is close to 0 which leads to almost no information propagation between nodes of different types. Please note that this method only applies to GMN since GEM's propagation component does not include a cross-graph attention matching. Regarding the taxonomy of informed machine learning (see Sect.~\ref{sec:Foundations:subsec:InformedML}), the source of the knowledge is given by the similarity definitions for semantic graphs. It is represented in form of mapping constraints and directly integrated into the hypothesis set. In contrast to the first extension, there is no change of the input data required.

\section{Experimental Evaluation}
\label{sec:Evaluation}

\begin{table*}[tb]
    \centering
    \includegraphics[width=0.95\textwidth]{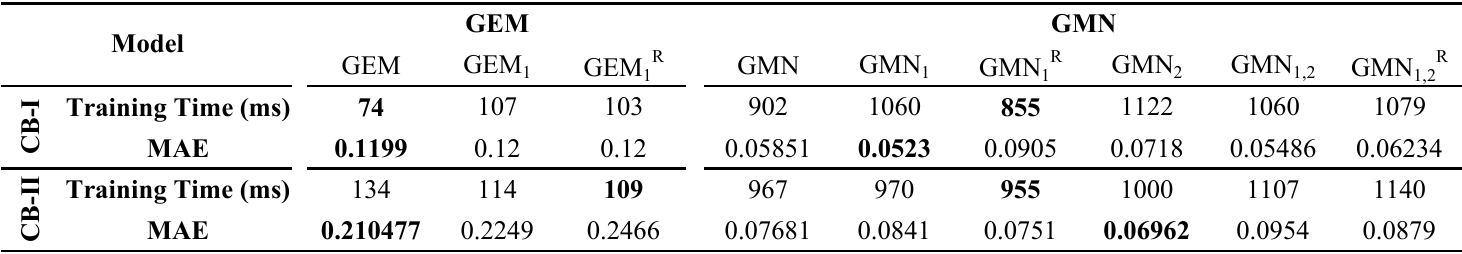}
    \caption[]{Evaluation Results}
    \label{tab:Eval}
\end{table*}

We evaluate our extended GNNs by comparing them to the stock models of GEM and GMN (see Sect.~\ref{sec:Concept:subsec:GNN}). In addition, we compare the effects of different combinations of the two extensions which are indicated by subscripts and superscripts in our terminology: GEM\textsubscript{1}, for instance, is GEM extended with the first extension. Furthermore, for the first extension, a superscript denotes whether the propagation layers are reused or not, \eg, GEM\textsubscript{1}\textsuperscript{R} for reused layers (see Sect.~\ref{sec:Concept:subsec:Extension1} for the explanation of layer reuse). The second extension is also marked with a subscript, \eg, GMN\textsubscript{2}. Combinations of both extensions are marked with both numbers as a subscript, \eg, GMN\textsubscript{1,2}. This gives a total of nine evaluated variants (see the columns in Tab.~\ref{tab:Eval} for the entire enumeration).

We measure the \textit{Mean Average Error} (MAE) of the predicted similarities as an indication of how well the model learns to assess pairwise case similarities, which correlates with suitable retrieval quality. In addition, we look at the influence of the extensions on the performance of each model. We take the training time as a representative measure since performance changes according to training time can be commonly transferred to other indicators, \eg, test time, validation time or inference latency \cite{Coleman.2019_DAWNBench}. Additionally, we report only the relative changes rather than the absolute time values in order to allow a direct comparison for this evaluation. We investigate the following hypotheses:
\begin{enumerate}[leftmargin=1cm]
    \item[\textbf{H1}] \hypertarget{H1}{Integrating} domain knowledge into GEM and GMN can improve the quality of similarity assessment \wrt MAE.
    \item[\textbf{H2}] \hypertarget{H2}{The} integration of domain knowledge into GEM and GMN can decrease training time.
\end{enumerate}
The hypotheses also aim at investigating a possible trade-off between quality improvements (\hyperlink{H1}{H1}) and increased training times (\hyperlink{H2}{H2}). Unless the extensions decrease both the MAE and the training time, it has to be discussed which applications benefit from which approach.

\subsection{Experimental Setup}
\label{sec:Evaluation:subsec:Setup}

We perform our experiments with two case bases from different domains. The cooking workflows (CB-I) are derived from 40 manually-modeled cooking recipes that are extended to 800 workflows by generalizing and specializing ingredients and cooking steps \cite{Muller.2018_WMA}, resulting in 660 training cases, 60 validation cases, and 80 test cases. The workflows of the data mining domain (CB-II) are built from sample processes that are delivered with RapidMiner (see \cite{Zeyen.2019_AdaptationScientificWorkflows} for more details), resulting in 509 training cases, 40 validation cases, and 60 test cases. The training cases are used as training input for the neural networks while the validation cases are used to monitor the training process and to optimize hyperparameter values. Hyperparameter tuning is performed individually per domain with the two stock models of GEM and GMN. The extended models then use the same hyperparameter settings as the associated stock model, \eg, GEM\textsubscript{1}\textsuperscript{R} uses the same hyperparameter configuration as GEM in the same domain. The measurements of MAE compare the predictions of the GNNs with the ground-truth similarity values according to our semantic graph similarity measure (see Sect.~\ref{sec:Foundations:subsec:SemanticGraphs} and \cite{Hoffmann.2020_GraphEmbedding} for more details). Those values are determined by computing pairwise similarities for all testing cases and averaging the values to a single number. Additionally, the average time for a single training iteration with 64 workflow pairs is measured to provide an insight of the variants' effects on training time. The machine that is used for training and testing computations is a PC with an Intel i7 6700 CPU (4 cores, 8 threads), an NVIDIA GTX 3070 GPU, and 48 GB of RAM, running Windows 10 64-bit.

\subsection{Experimental Results}
\label{sec:Evaluation:subsec:Results}

The evaluation results are depicted in Tab. \ref{tab:Eval}. The table shows the MAE and the average training iteration time (in milliseconds) of all evaluated models and for both case bases. The lowest MAE and training time values are highlighted in bold font, grouped by base model (first line) and domain (rotated text on the left).

The quality of the models \wrt MAE shows no significant influence of the extensions on the GEM model. Both for CB-I and CB-II, the best MAE is provided by the stock GEM model. While the extended models of CB-I are on par with the stock model, the extended models of CB-II show a decreased quality with a maximum reduction of approx. 17\% for GEM\textsubscript{1}\textsuperscript{R}. The effectiveness of the extended GMN variants is different. The GMN stock model is not the best performing model for either of the case bases. In fact, the extensions lead to a maximum MAE decrease of approx. 12\% for CB-I and of approx. 10\% for CB-II. But not all extensions decreased the MAE and the success of one extension is not consistent for both case bases. For instance, whereas GEM\textsubscript{1} shows the best overall MAE for CB-I it does not lead to a decrease when applied to CB-II. The same results can be observed for GEM\textsubscript{2} when applied to CB-II.

The time values show different effects of the extensions for either of the case bases and the base models. Integrating the first extension into GEM results in time increases for CB-I with a maximum of approx. 44\%. For GEM and CB-II, however, the first extension brings performance benefits of 17 - 23\%. The extended variants of GMN show a similar picture. The first extension has similar run times compared to the stock GMN model. Only GMN\textsubscript{1} for CB-I sticks out with an increase of approx. 17\%. The second extension only has a large impact on performance for CB-I, \ie, a maximum of 24\% slower than the stock model. For CB-II, the combination of both extensions seems to increase the training times. It is also apparent that GMN\textsubscript{1}\textsuperscript{R} decreased the training time for both case bases compared to the stock model and achieves the lowest values of the GMN variants.

\subsection{Discussion}
\label{sec:Evaluation:subsec:Discussion}

The most noticeable observation of this experiment is the inconsistency of the effects that different extensions have on different domains. Compared to the stock models, those effects range from a decrease in training time and MAE, \eg, GMN\textsubscript{1}\textsuperscript{R}, to an increase in training time and MAE, \eg, GMN\textsubscript{2}. This leads to the preliminary result of this experiment: Whether an extension method is beneficial for the underlying GNN is highly dependent on the method itself and the target domain. Individual testing of different methods and subsequent tuning is inevitable. The effect of the first extension on GEM is a good example in which way more general knowledge about the case representation can lead to different results in different domains. This extension significantly increases the training time for CB-I and decreases it for CB-II, compared to the stock models. This is reasonable as CB-II tends to have larger semantic descriptions and graphs that are more affected by the slow, sequential computations of RNNs which process the semantic descriptions in the stock models (see Sect.~\ref{sec:Concept:subsec:Extension1}). The graphs and semantic descriptions of CB-I are smaller than those of CB-II and, thus, show an opposing effect where the information propagation in the GNNs is more expensive than the RNN computations. The example also shows the trade-off between quality and time: In most cases, a reduction of the MAE results in an increase of time. This means that the underlying domain should be analyzed for its requirements regarding these two aspects and appropriate benchmarks should be conducted. Considering the discussed results, \hyperlink{H1}{H1} can be accepted for GMN due to multiple retriever variants that improve quality, compared to the stock models. \hyperlink{H2}{H2} can also be accepted for GMN due to GMN\textsubscript{1}\textsuperscript{R} which leads to a decreased training time. The experiments show less indications of positive effects for GEM. The first extension fails at improving the quality for both CB-I and CB-II. Thus, \hyperlink{H1}{H1} is rejected for GEM. \hyperlink{H2}{H2} can be partly accepted due to the positive effects of the first extension on the training time for CB-II.

\section{Conclusion and Future Work}
\label{sec:ConclusionAndFutureWork}

This paper examines the potential of including domain knowledge of \textit{Case-Based Reasoning} (CBR) applications into \textit{Deep Learning} (DL) models for learning similarities between semantic graphs. We present two different methods for integration into two \textit{Graph Neural Networks}. The first method aims at representing the semantic annotations of nodes and edges in a tree structure that enables the usage of message-passing neural networks for processing. The tree structure reflects the implicit relations of the data which is composed of composite and atomic data types. The second method introduces constraints to the attention-based matching procedure of the GNN, based on domain knowledge about legal node and edge mappings in the similarity assessment procedure. This method guides the DL model to use very low attention for non-legal mappings and higher attention for legal mappings. Both extensions are part of an experimental evaluation, focusing on changes in retrieval quality and training time. The results show that the extensions are able to improve quality and to reduce training time when being compared to the stock models. However, the effects are not consistent across different domains and different models.

A focus of future work should be on conducting a more comprehensive evaluation of the approaches from this paper in order to measure the influence of knowledge integration for more domains. It has the goal of verifying the results and finding more integration opportunities in the research field of CBR. Additionally, the ideas of informed machine learning \cite{VonRueden.2019_InformedMachineLearning} can be used to provide a more detailed overview of the challenges and opportunities when being applied to CBR. Furthermore, the approaches from this paper can especially be a chance for applications that deal with case adaptation and DL, \eg, \cite{Liao.2018_MLCaseAdaptation,Leake.2021_AdaptationNeuralNetworks}. Since adaptation is usually a knowledge-intensive process, our approaches can help by providing expressive learning capabilities combined with possibilities of domain knowledge integration.


\bibliographystyle{named}
\bibliography{bibliography}

\end{document}